# Anytime Marginal Maximum a Posteriori Inference

**Denis Deratani Mauá**  DENIS@IDSIA.CH
IDSIA, Galleria 2, Manno, Switzerland, CH 6928

**Cassio Polpo de Campos**  CASSIO@IDSIA.CH
IDSIA, Galleria 2, Manno, Switzerland, CH 6928

## Abstract

This paper presents a new anytime algorithm for the marginal MAP problem in graphical models of bounded treewidth. We show asymptotic convergence and theoretical error bounds for any fixed step. Experiments show that it compares well to a state-of-the-art systematic search algorithm.

## 1. Introduction

The maximum a posteriori (MAP) assignment problem consists in finding an assignment that maximizes the posterior probability of a given set of variables. To facilitate modeling, the model often includes latent variables that are neither to be maximized nor observed, but marginalized. It is this more general form of the problem (a.k.a. partial or marginal MAP) that we tackle in this paper. Moreover, we assume that the probability distribution is represented as a discrete *graphical model*, which allows for compactness.

Computationally, this is a very hard problem. It is NP$^{PP}$-hard even if all variables are binary, and NP-hard if either the underlying graph has bounded treewidth or there are no latent variables (Park & Darwiche, 2004). Also producing a provably good approximate solution is NP-hard, even if the treewidth of the underlying graph is bounded (Park & Darwiche, 2004). A positive result has recently been given by de Campos (2011), which derived a fully polynomial-time approximation scheme when both treewidth and number of states per variable are bounded.

MAP assignment problems can be seen as a composition of two different tasks: the computation of marginal probabilities and the combinatorial search over assignments. The former is responsible for evaluating the quality of a candidate assignment produced by the latter. When the marginal probability inference is tractable, standard combinatorial search approaches such as branch-and-bound for exact solutions and local search for approximate results have been used (Park & Darwiche, 2003; Yuan et al., 2004). When it is hard, researchers have resorted to loopy belief propagation algorithms (Liu & Ihler, 2011; Jiang et al., 2011) and factor decomposition (Meek & Wexler, 2011).

In this paper, we present a new anytime algorithm to perform marginal MAP inference in graphical models of bounded treewidth. The algorithm implements a clique-tree propagation scheme that propagates sets of factors instead of single factors. Efficiency is achieved by verifying sub-optimality locally. We show empirically that the algorithm compares well to the systematic search algorithm of Park & Darwiche (2003). We derive theoretical bounds for the error produced by the algorithm within a given amount of computational resources (time and memory), and show that this error can be made arbitrarily small with enough resources.

## 2. Notation

A finite integer set $\{1, 2, \ldots, n\}$ is denoted by $[n]$. Random variables are represented by capital letters, e.g., $X, Y, Z$; real-valued functions by greek letters, e.g., $\phi$, $\psi$; sets by calligraphic letters, e.g., $\mathcal{I} = [3]$, $\mathcal{P} = \{\phi, \psi\}$, $\mathcal{S} = \{\mathcal{I}, \mathcal{P}\}$; vectors in boldface, e.g., $\mathbf{V} = (X, Y, Z)$. The number of elements in a set $\mathcal{X}$ is denoted by $|\mathcal{X}|$. We identify a variable with its sample space. Hence, the finite set of values a variable $X$ can assume is also denoted by $X$. Given a vector of variables $\mathbf{X} = (X_1, \ldots, X_n)$, we write $\mathbf{X} = X_1 \times \cdots \times X_n$ to denote the space of configurations or assignments of the variables in $\mathbf{X}$, where $\times$ denotes the Cartesian product. We also identify a vector of variables to its joint sample space, so that the notation $\mathbf{x} \in \mathbf{X}$ is well-defined, and $|\mathbf{X}|$ denotes the number of assignments $\mathbf{x}$ to **X** and not the number of variables in the vector. For $\mathbf{X}=(X_1,\ldots,X_n)$ and $\mathcal{I} \subseteq [n]$, the notation $\mathbf{X}_\mathcal{I}$ denotes the vector $(X_i)_{i\in\mathcal{I}}$. We write $\mathbf{x}_\mathcal{I}$ to denote the vector $(x_i)_{i\in\mathcal{I}}$ obtained by projecting $\mathbf{x} \in \mathbf{X}$ onto $\mathbf{X}_\mathcal{I}$.

A *factor* $\phi$ over a vector of variables $\mathbf{X}=(X_1,\ldots,X_n)$ is a $|\mathbf{X}|$-dimensional vector of non-negative real values. The value of the factor corresponding to a particular assignment $\mathbf{x} \in \mathbf{X}$ is denoted by $\phi(\mathbf{x})$. Given an assignment $\mathbf{z}$ for a vector of variables $\mathbf{X}$, the *indicator factor* $\delta_\mathbf{z}$ assigns value one for $\mathbf{x}=\mathbf{z}$ and zero for all others. Product and sum-marginalization of factors are defined as usual: $(\phi\psi)(\mathbf{x}) = \phi(\mathbf{x}_\mathcal{I})\psi(\mathbf{x}_\mathcal{J})$, for $\mathbf{X}=(X_1,\ldots,X_n), \mathcal{I}\cup\mathcal{J}=[n]$, $\phi$ defined over $\mathbf{X}_\mathcal{I}$ and $\psi$ defined over $\mathbf{X}_\mathcal{J}$; $\left(\sum_{\mathbf{X}_\mathcal{I}} \phi\right)(\mathbf{y}) = \sum_{\mathbf{x}\in\mathbf{X}} \phi(\mathbf{x})\delta_\mathbf{y}(\mathbf{x}_\mathcal{I})$ for $\phi$ defined over $\mathbf{X}=(X_1,\ldots,X_n), \mathcal{I} \subseteq [n]$ and $\mathbf{y} \in \mathbf{X}_\mathcal{I}$.

Given a tree $\mathcal{T}$ over $[n]$ and a root node $r \in [n]$, we say that a node $p$ is the parent of a neighboring node $i$ if $p$ is closer to $r$ than $i$, in which case, we say that $i$ is a child of $p$. The parent and the children of a node $i$ are denoted by $pa(i)$ and $ch(i)$, respectively. The set of descendants of $i$ (i.e., its children, the children of its children, and so on) is denoted by $de(i)$. Nodes containing no children are called leaves, and nodes containing at least one child are called internal.

## 3. Graphical Models

Let $\mathbf{X} = (X_1,\ldots,X_n)$ be a vector of discrete random variables, $\mathcal{J}_1,\ldots,\mathcal{J}_m$ be a collection of index sets satisfying $\mathcal{J}_1 \cup \cdots \cup \mathcal{J}_m = [n]$, and $\mathcal{P} = \{\phi_1,\ldots,\phi_m\}$ be a set of factors over vectors $\mathbf{X}_{\mathcal{J}_1},\ldots,\mathbf{X}_{\mathcal{J}_m}$, respectively. We call $\mathcal{P}$ a *graphical model* if it specifies a joint probability distribution over assignments $\mathbf{x} \in \mathbf{X}$ by

$$\Pr(\mathbf{X}=\mathbf{x}) = \frac{1}{Z} \prod_{i\in[m]} \phi_i(\mathbf{x}_{\mathcal{J}_i}),$$

where $Z = \sum_\mathbf{X} \prod_{\phi\in\mathcal{P}} \phi$ is a normalizing constant known as the *partition function*. The graph in the left-hand side of Figure 1 depicts a graphical model.

Let $\mathcal{D}$ and $\mathcal{H}$ be index sets partitioning the variables into decision and latent variables, respectively. The *MAP assignment problem* consists in finding

$$\begin{aligned}
\mathbf{d}^* &= \operatorname*{argmax}_{\mathbf{d}\in\mathbf{X}_\mathcal{D}} \Pr(\mathbf{X}_\mathcal{D}=\mathbf{d}) \\
&= \operatorname*{argmax}_{\mathbf{d}\in\mathbf{X}_\mathcal{D}} \sum_{\mathbf{h}\in\mathbf{X}_\mathcal{H}} \Pr(\mathbf{X}_\mathcal{D}=\mathbf{d}, \mathbf{X}_\mathcal{H}=\mathbf{h}) \\
&= \operatorname*{argmax}_{\mathbf{d}\in\mathbf{X}_\mathcal{D}} \sum_\mathbf{X} \prod_{i\in[m]} \phi_i \prod_{j\in\mathcal{D}} \delta_{\mathbf{d}_j}. \quad (1)
\end{aligned}$$

For each fixed assignment $\mathbf{d}$, we can represent the factorization in (1) by a new graphical model $\mathcal{P}_\mathbf{d} = \mathcal{P} \cup$ $\bigcup_{j\in\mathcal{D}}\{\delta_{\mathbf{d}_j}\}$. The partition function of this new model satisfies $Z_\mathbf{d} = \sum_\mathbf{X} \prod_{i\in[m]} \phi_i \prod_{j\in\mathcal{D}} \delta_{\mathbf{d}_j}$. This way, we can re-state the MAP assignment problem as a search over graphical models $\mathcal{P}_\mathbf{d}$. Assume without loss of generality that $\mathcal{D}=\{1,\ldots,d\}$ and $\mathcal{H}=\{d+1,\ldots,n\}$, and define $\mathcal{K}_i = \{\phi_i\}$ for $i = 1,\ldots,m$, and $\mathcal{K}_{i+m} = \{\delta_{x_i} : x_i \in X_i\}$ for each decision $i \in \mathcal{D}$. Each combination of factors $\phi_1,\ldots,\phi_{m+d}$ from sets $\mathcal{K}_1,\ldots,\mathcal{K}_{m+d}$, respectively, specifies the graphical model $\mathcal{P}_\mathbf{d}$ corresponding to an assignment $\mathbf{d}$. Let $\mathcal{M} = \{\{\phi_1,\ldots,\phi_{m+d}\} : \phi_i \in \mathcal{K}_i\}$ denote all graphical models obtained in such a way. Finding a MAP assignment is equivalent to finding a graphical model $\mathcal{P}^* = \operatorname{argmax}_{\mathcal{P}\in\mathcal{M}} \sum_\mathbf{X} \prod_{\phi\in\mathcal{P}} \phi$. An assignment $\mathbf{d}^*$ is a MAP assignment iff it satisfies $\mathbf{d}^* = \operatorname{argmax}_{\mathbf{d}\in\mathbf{X}_\mathcal{D}} \prod_{i=m+1}^{m+d} \phi_i(\mathbf{d}_i)$ for some optimal $\mathcal{P}^*$.

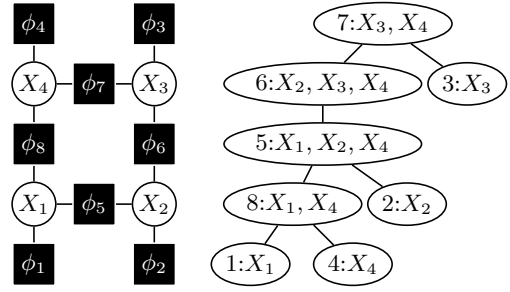

*Figure 1.* A simple graphical model represented as a factor graph (on the left) and a suitable clique tree (on the right). Each factor $\phi_i$, $i = 1,\ldots,8$, is defined over the variables in neighboring nodes and assigned to the clique tree node $i$.

**Example 1.** *Consider the graphical model in Figure 1, and assume that variables are binary, $\mathcal{D}=\{1,2\}$ and $\mathcal{H}=\{3,4\}$. We denote the values a binary variable $X_i$ can assume by $x_i$ and $\tilde{x}_i$, and reformulate this MAP assignment problem as a search over graphical models as follows. Let $\mathcal{K}_1 = \{\phi_1\}$, $\mathcal{K}_2 = \{\phi_2\}$, $\mathcal{K}_3 = \{\phi_3\}$, $\mathcal{K}_4 = \{\phi_4\}$, $\mathcal{K}_5 = \{\phi_5\}$, $\mathcal{K}_6 = \{\phi_6\}$, $\mathcal{K}_7 = \{\phi_7\}$, $\mathcal{K}_8 = \{\phi_8\}$, $\mathcal{K}_9 = \{\delta_{x_1},\delta_{\tilde{x}_1}\}$ and $\mathcal{K}_{10} = \{\delta_{x_2},\delta_{\tilde{x}_2}\}$. Each combination of factors $\phi_1,\ldots,\phi_{10} \in \mathcal{K}_1,\ldots,\mathcal{K}_{10}$ corresponds to the graphical model induced by the assignment $\mathbf{d} = \operatorname{argmax}_\mathbf{x} \phi_9(\mathbf{x}_1)\phi_{10}(\mathbf{x}_2)$. Suppose that $\mathcal{P}^* = \{\phi_1,\ldots,\phi_8,\delta_{x_1},\delta_{\tilde{x}_2}\}$ is a solution to $\operatorname{argmax}_{\mathcal{P}\in\mathcal{M}} \sum_{X_1,X_2,X_3,X_4} \prod_{i\in[10]} \phi_i$. Then $\mathbf{d}^* = (x_1, \tilde{x}_2) = \operatorname{argmax}_\mathbf{d} \delta_{x_1}(\mathbf{d}_1)\delta_{\tilde{x}_2}(\mathbf{d}_2)$ is a MAP assignment.*

## 4. Clique-Tree Computation

Let $\mathcal{T}$ be a tree over $[m]$, $\mathcal{I}_1,\ldots,\mathcal{I}_m$ be a collection of index sets satisfying $\mathcal{I}_1 \cup \cdots \cup \mathcal{I}_m = [n]$ for some positive integer $n$. We call $\mathcal{T}$ a *clique tree* if for $i = 1,\ldots,n$ the subgraph obtained by removing from $\mathcal{T}$ all nodes $j$ such that $i \notin \mathcal{I}_j$ remains a tree. Clique trees are so called because the index sets usually represent the cliques in the triangulated underlying graph of a graphical model. Let $\mathcal{P}$ be a graphical model whose factors $\phi_1, \ldots, \phi_k$ are defined over sets $\mathbf{X}_{\mathcal{J}_1}, \ldots, \mathbf{X}_{\mathcal{J}_k}$, respectively, and $\mathcal{J}_1 \cup \cdots \cup \mathcal{J}_k = [n]$. We say that $\mathcal{T}$ is a clique tree for $\mathcal{P}$ if for $i = 1, \ldots, k$ there is $1 \leq j \leq m$ such that $\mathcal{J}_i \subseteq \mathcal{I}_j$. In the following, we assume for ease of exposition and without loss of generality that if $\mathcal{T}$ is a clique tree for $\mathcal{P}$ then $m = k$ and $\mathcal{J}_i \subseteq I_i$ for all $i$, which allows us to unambiguously associate each factor $\phi_i$ to the clique tree node $i$. The tree on the right-hand side in Figure 1 is a clique tree for the graphical model on the left.

The *width* of a clique tree is the cardinality of the largest index set minus one. For example, the width of the tree in Figure 1 is two. Since the complexity of algorithms that operate on clique trees is (at least) exponential in the tree width, one usually seeks to obtain a clique tree of low width. Finding a minimum-width clique tree for a given graphical model is an NP-hard problem, and one usually resorts to heuristics to obtain low-width trees.

The basic computation scheme with clique trees is the FACTOR-ELIMINATION procedure in Algorithm 1, which computes the partition function of a graphical model $\mathcal{P} = \{\phi_1, \ldots, \phi_m\}$ associated to a clique tree $\mathcal{T}$ over $[m]$.[1] In the algorithm, we assume that each factor $\phi_i$ is assigned to node $i$ in the clique tree (hence its associated index set $\mathcal{J}_i \subseteq \mathcal{I}_i$). In a nutshell, the algorithm roots the tree in an arbitrary node $r$, and then propagates messages from the leaves towards the root. For ease of exposition, we assume in line 5 that $\mathcal{I}_{pa(r)} = \emptyset$. The propagation of messages halts when the root receives a message from every child, in which case the partition function is obtained by $Z = \mu_r$. The algorithm runs in $O(msw^{+1})$ time, where $s = \max_i |X_i|$ is the maximum number of values a variable in the model can assume, and $w = \max_i |\mathcal{I}_i| - 1$ is the width of the clique tree. Thus when the width $w$ is bounded, the computations take polynomial time.

Let $h(i) = \bigcup_{j \in de(i) \cup \{i\}} \mathcal{I}_j \setminus \mathcal{I}_{pa(i)}$. It can be shown that for $i = 1, \ldots, m$ the factor $\mu_i$ satisfies $\mu_i = \sum_{\mathbf{X}_{h(i)}} \phi_i \prod_{j \in de(i)} \phi_j$ (Koller & Friedman, 2009). Since $h(r) = [n]$ by definition of clique trees, the correctness of the computations follows easily by applying this result to the root: $Z = \mu_r = \sum_{\mathbf{X}} \prod_{i \in [m]} \phi_i$. Hence, we can evaluate the quality of a candidate solution $\mathbf{d}$ to the MAP assignment problem by building a clique tree $\mathcal{T}$ for the corresponding graphical model $\mathcal{P}_\mathbf{d}$ and then running FACTOR-ELIMINATION, which produces

---

[1]The name COLLECT algorithm has also been used to describe the same algorithm.

---

**Algorithm 1** FACTOR-ELIMINATION
**Require:** A clique tree $\mathcal{T}$ over a graphical model $\mathcal{P}$
**Ensure:** $Z = \sum_\mathbf{X} \prod_{\phi \in \mathcal{P}} \phi$
1: select a node $r$ as root
2: label all nodes as inactive
3: **while** there is an inactive node $i$ **do**
4:    select an inactive node $i$ with all children active
5:    compute $\mu_i = \sum_{\mathbf{X}_{\mathcal{I}_i \setminus \mathcal{I}_{pa(i)}}} \phi_i \prod_{j \in ch(i)} \mu_j$
6:    label $i$ as active
7: **end while**
8: $Z = \mu_r$

---

$Z_\mathbf{d} = \sum_\mathbf{X} \prod_{\phi \in \mathcal{P}_\mathbf{d}} \phi$. Note that the same clique tree can be used to evaluate different candidates.

**Example 2.** *Consider the graphical model and clique tree in Figure 1 and assume again that variables are binary $\mathcal{D} = \{1, 2\}$ and $\mathcal{H} = \{3, 4\}$. We can evaluate the assignment $\mathbf{d} = (x_1, \tilde{x}_2)$ to $(X_1, X_2)$ by replacing $\phi_1$ and $\phi_2$ with $\phi_1' = \phi_1 \delta_{x_1}$ and $\phi_2' = \phi_2 \delta_{\tilde{x}_2}$, respectively, and then running FACTOR-ELIMINATION, which obtains $Z_{(x_1, \tilde{x}_2)} = \sum_{X_1, X_2, X_3, X_4} \prod_{i=1}^8 \phi_i \delta_{x_1} \delta_{\tilde{x}_2} \propto \Pr(X_1 = x_1, X_2 = \tilde{x}_2)$.*

The algorithm can be straightforwardly modified to find a MAP assignment when there are no latent variables (i.e., when $\mathcal{H} = \emptyset$) by substituting sums with maximizations in the computation of factors $\mu_i$ (Koller & Friedman, 2009). This naturally suggests an approach to the computation of MAP assignments in the presence of latent variables (i.e., when $\mathcal{H} \neq \emptyset$), which consists in redefining the factors $\mu_i$ so that latent variables are summed out while decision variables are maximized. A FACTOR-MAX-ELIMINATION version of the algorithm thus obtains factors $\mu_i = \max_{\mathbf{X}_{\mathcal{D}_i}} \sum_{\mathbf{X}_{\mathcal{H}_i}} \phi_i \prod_{j \in ch(i)} \mu_j$, where $\mathcal{D}_i = (\mathcal{I}_i \cap \mathcal{D}) \setminus \mathcal{I}_{pa(i)}$ and $\mathcal{H}_i = (\mathcal{I}_i \cap \mathcal{H}) \setminus \mathcal{I}_{pa(i)}$. Variants of this procedure have recently been justified as an approximation by variational inference (Liu & Ihler, 2011; Jiang et al., 2011). These approaches retain the efficiency of message-passing algorithms, but produce only an upper bound to the real value, unless the root node $r$ contains all decision variables. Enforcing the clique tree to contain a node over all decision variables results in an exponential complexity in the number of decision variables (Park & Darwiche, 2004), unless the factors in the root node are factorized (Meek & Wexler, 2011).

Another simple but often effective approach to the MAP assignment problem is to perform a search over the space of assignments, and to use FACTOR-ELIMINATION to evaluate candidate solutions. An upper bound for any partial assignment can be obtained by running FACTOR-MAX-ELIMINATION, which poten-

**Algorithm 2** FACTOR-SET-ELIMINATION
**Require:** A clique tree $\mathcal{T}$ over the sets of factors $\mathcal{K}_1, \ldots, \mathcal{K}_m$ and positive integers $k_1, \ldots, k_m$
**Ensure:** $Z_l \leq Z^* \leq Z_u$
1: select a node $r$ as root and let $\sigma$ be an empty dictionary
2: **for all** leaf node $i$ **do**
3:     let $\mathcal{M}_i$ be an empty set
4:     **for all** $\phi_i \in \mathcal{K}_i$ **do**
5:         add $\mu_i = \sum_{\mathbf{X}_{\mathcal{I}_i \setminus \mathcal{I}_{pa(i)}}} \phi_i$ to $\mathcal{M}_i$
6:         set $\sigma(\mu_i) \leftarrow \mu_i$
7:     **end for**
8:     $\mathcal{L}_i = \text{prune}(\mathcal{M}_i, \sigma_i, k_i)$
9: **end for**
10: label leaves as active and internal nodes as inactive
11: **while** there is an inactive node **do**
12:     select an inactive node $i$ whose children are all active
13:     let $\mathcal{M}_i$ be empty sets
14:     **for all** $\phi_i \in \mathcal{K}_i, \mu_j \in \mathcal{L}_j, j \in ch(i)$ **do**
15:         add $\mu_i = \sum_{\mathbf{X}_{\mathcal{I}_i \setminus \mathcal{I}_{pa(i)}}} \phi_i \prod_{j \in ch(i)} \mu_j$ to $\mathcal{M}_i$
16:         set $\sigma(\mu_i) \leftarrow \sum_{\mathbf{X}_{\mathcal{I}_i \setminus \mathcal{I}_{pa(i)}}} \phi_i \prod_{j \in ch(i)} \sigma(\mu_j)$
17:     **end for**
18:     $\mathcal{L}_i = \text{prune}(\mathcal{M}_i, \sigma, k_i)$
19:     label $i$ as active
20: **end while**
21: $Z_l = \max\{\mu_r : \mu_r \in \mathcal{L}_r\}$
22: $Z_u = \max\{\sigma(\mu_r) : \mu_r \in \mathcal{L}_r\}$

---

tially narrows the search space. The algorithm of Park & Darwiche (2003), against which we compare the algorithm we devise here, builds on this idea.

## 5. Propagating Sets

Recall from the previous section that we can compare the quality of different candidate solutions to the MAP assignment problem by running FACTOR-ELIMINATION with the same clique tree structure but different indicator factors. More generally, let $\mathcal{K}_1, \ldots, \mathcal{K}_m$ be a collection of sets of factors such that each $\mathcal{P} = \{\phi_1, \ldots, \phi_m\}$ obtained by selecting a factor $\phi_i$ from $\mathcal{K}_i$, $i = 1, \ldots, m$, is a graphical model. Let $\mathcal{P}$ be a graphical model obtained in this way, and let $\mathcal{T}$ be a clique tree for this model. Then $\mathcal{T}$ is also a clique tree for any other graphical model induced by $\mathcal{K}_1, \ldots, \mathcal{K}_m$. This insight is the base of the FACTOR-SET-ELIMINATION procedure in Algorithm 2, which performs a search over the space of assignments while it propagates sets of factors over the clique tree.

The algorithm resembles FACTOR-ELIMINATION, but instead of propagating factors $\mu_i$, it propagates sets of factors $\mathcal{L}_i \subseteq \mathcal{M}_i = \{\sum_{\mathbf{X}_{\mathcal{I}_i \setminus \mathcal{I}_{pa(i)}}} \phi_i \prod_{j \in ch(i)} \mu_j : \phi_i \in \mathcal{K}_i, \mu_j \in \mathcal{L}_j\}$. The elements $\sigma(\mu_i)$ obtained in lines 6 and 16 are local upper bounds which we discuss later on. The pruning operations in lines 8 and 18 return a subset $\mathcal{L}_i \subseteq \mathcal{M}_i$ of cardinality $k_i$ and recompute the upper bounds $\sigma(\mu_i)$ to account for the discarded elements. So, if $k_i \geq |\mathcal{M}_i|$, then the pruning operation returns $\mathcal{L}_i = \mathcal{M}_i$. The algorithm outputs lower and upper bounds $Z_l$ and $Z_u$, respectively, to the maximum partition function $Z^* = \max\{\sum_{\mathbf{X}} \prod_{i \in [m]} \phi_i : \phi_i \in \mathcal{K}_i\}$ of a graphical model induced by the sets in the input. The following result shows the correspondence of factors $\mu_i$ computed by this algorithm to those computed with FACTOR-ELIMINATION.

**Theorem 1.** *For $i = 1, \ldots, m$, any $\mu_i \in \mathcal{L}_i$ satisfies $\mu_i = \sum_{\mathbf{X}_{h(i)}} \phi_i \prod_{j \in de(i)} \phi_j$ for some combination of $\phi_i \in \mathcal{K}_i$ and $\phi_j \in \mathcal{K}_j$ for all $j \in de(i)$.*

*Proof.* First, note that the definition of $\mu_i$ in FACTOR-SET-ELIMINATION is identical to the definition in FACTOR-ELIMINATION. Assume the pruning operations are not performed, that is, that $\text{prune}(\mathcal{M}_i, \sigma_i, k_i)$ returns $\mathcal{M}_i$. Then it is not difficult to see that $\mu_i$ matches the computation in FACTOR-ELIMINATION for some graphical model induced by $\mathcal{K}_1, \ldots, \mathcal{K}_m$. But since the pruning operation returns a subset of $\mathcal{M}_i$, this holds also for any $\mu_i \in \mathcal{L}_i$. □

The following result follows immediately from the above theorem.

**Corollary 1.** $Z_l = \sum_{\mathbf{X}} \prod_{i \in [m]} \phi_i$ *for some combination of factors $(\phi_1, \ldots, \phi_m) \in \mathcal{K}_1 \times \cdots \times \mathcal{K}_m$.*

If the algorithm is run with factor sets $\mathcal{K}_1, \ldots, \mathcal{K}_m$ that induce graphical models corresponding to different assignments to decision variables as explained in Section 3, the numbers $Z_l$ and $Z_u$ returned are lower and upper bounds for the MAP assignment probability $Z^* = \max_{\mathbf{d}} \Pr(\mathbf{X}_\mathcal{D} = \mathbf{d})$. In fact, if $k_i = |\mathcal{M}_i|$ for all $i = 1, \ldots, m$, the algorithm is equivalent to an exhaustive search over the space of assignments, and thus returns $Z_l = Z^*$. Moreover, the value of $Z_l$ is actually achieved by some assignment, and hence denotes the value of a feasible solution. The assignment corresponding to $Z_l$ can be obtained by tracking back the indicator factors $\delta_i$, $i \in \mathcal{D}$, that were propagated to generate the number $\mu_r = Z_l$.

The complexity of the algorithm is determined by the number of additions and multiplications needed to compute each factor $\mu_i$ in a set $\mathcal{M}_i$ plus the complexity of the pruning operation. Similarly to FACTOR-ELIMINATION, the complexity of computing each $\mu_i$ is $O(ms^{w+1})$. Let $k$ be the maximum of $k_1, \ldots, k_m$ and $|\mathcal{K}_1|, \ldots, |\mathcal{K}_m|$. By design, each set $\mathcal{M}_i$ contains $|\mathcal{K}_i| \prod_{j \in ch(i)} |\mathcal{L}_j| = |\mathcal{K}_i| \prod_{j \in ch(i)} k_j \leq k^c$ elements, where $c$ is the maximum number of neighbors of a node. Hence, the algorithm runs in $O(k^c m s^w)$. If the clique tree given as input contains a bounded number of children for each node and bounded width, the algorithm runs in time polynomial in the inputs $k_1, \ldots, k_m$ and $\mathcal{K}_1, \ldots, \mathcal{K}_m$. Note that for any given any graphical model of bounded treewidth we can obtain a clique tree of bounded width and bounded number of children per node (e.g., a binary clique tree).

### 5.1. Pruning

The pruning operations are responsible for reducing the size of the propagated sets, enabling efficient inference. The trade-off between the quality of the solution and the computation time is determined by the parameters $k_1, \ldots, k_m$ in the input. In the following, we discuss how the pruning operations are implemented.

Consider a set of factors $\mu_i^{(1)}, \ldots, \mu_i^{(k)}$ which we wish to discard to reduce the size of a set $\mathcal{M}_i$ produced during FACTOR-SET-ELIMINATION. Our first insight is that convex combinations can be safely removed, as they are certainly outperformed by some extrema.

A factor $\mu_i^{(1)}$ is a *convex combination* of factors $\mu_i^{(2)}$ and $\mu_i^{(3)}$ if there is a real $0 \leq \lambda \leq 1$ such that $\mu_i^{(1)} = \lambda \mu_i^{(2)} + (1 - \lambda) \mu_i^{(3)}$. Given a set of factors $\mathcal{M}_i$, we say that $\mu_i \in \mathcal{M}_i$ is an *extreme* if it is not a convex combination of any two other elements in the set. Non-extreme factors can be safely removed from $\mathcal{M}_i$, as the following result shows.

**Theorem 2.** *Let $\mu_i^{(1)}$, $\mu_i^{(2)}$ and $\mu_i^{(3)}$ be three different factors in a set $\mathcal{M}_i$ such that $\mu_i^{(1)}$ is a convex combination of $\mu_i^{(2)}$ and $\mu_i^{(3)}$. Then any solution value $\mu_r^{(1)}$ different from $\mu_r^{(2)}$ and $\mu_r^{(3)}$, where $\mu_r^{(\ell)}$ is obtained by propagating $\mu_i^{(\ell)}$ up to the root, is not an optimal solution.*

*Proof.* Let $\mu_j^{(1)} = \sum_{X_{\mathcal{I}_j \setminus \mathcal{I}_p}} \phi_j \mu_i^{(1)} \prod_{k \in ch(j) \setminus \{i\}} \mu_k$, $\mu_j^{(2)} = \sum_{X_{\mathcal{I}_j \setminus \mathcal{I}_p}} \phi_j \mu_i^{(2)} \prod_{k \in ch(j) \setminus \{i\}} \mu_k$ and $\mu_j^{(3)} = \sum_{X_{\mathcal{I}_j \setminus \mathcal{I}_p}} \phi_j \mu_i^{(3)} \prod_{k \in ch(j) \setminus \{i\}} \mu_k$ be factors in $\mathcal{M}_j$, where $j = pa(i)$ and $p = pa(j)$. Then $\mu_j^{(1)}$ is a convex combination of $\mu_j^{(2)}$ and $\mu_j^{(3)}$. By induction in the nodes of the clique tree, we find that any number $\mu_r^{(1)} \in \mathcal{M}_r$ obtained by propagating $\mu_i^{(1)}$ up to the root is a convex combination of numbers $\mu_r^{(2)}$ and $\mu_r^{(3)}$ obtained by propagating $\mu_i^{(2)}$ and $\mu_i^{(3)}$, respectively, up to the root. Hence, $\mu_r^{(1)}$ is necessarily (strictly) less than $\max\{\mu_r^{(2)}, \mu_r^{(3)}\}$, which is less than or equal to the optimal solution $Z^*$. □

There is also another condition between factors which if verified allows us to safely discard a factor from $\mathcal{M}_i$. Let $\mu_i^{(1)}$ and $\mu_i^{(2)}$ be two factors in $\mathcal{M}_i$. We say that $\mu_i^{(2)}$ (weakly Pareto-)dominates $\mu_i^{(1)}$, and write $\mu_i^{(2)} \geq \mu_i^{(1)}$, if $\mu_i^{(2)}(\mathbf{x}) \geq \mu_i^{(1)}(\mathbf{x})$ for all $\mathbf{x} \in \mathbf{X}_{\mathcal{I}_i \cap \mathcal{I}_{pa(i)}}$. As the following result shows, we can safely remove dominated factors.

**Theorem 3.** *Let $\mu_i^{(1)}$ and $\mu_i^{(2)}$ be two different factors in a set $\mathcal{M}_i$ such that $\mu_i^{(2)} \geq \mu_i^{(1)}$. Then any solution $\mu_r^{(1)} \neq \mu_r^{(2)}$, where $\mu_r^{(\ell)}$ is obtained by propagating $\mu_i^{(\ell)}$ up to the root, is not an optimal solution.*

*Proof.* Let $\mu_j^{(1)} = \sum_{X_{\mathcal{I}_j \setminus \mathcal{I}_p}} \phi_j \mu_i^{(1)} \prod_{k \in ch(j) \setminus \{i\}} \mu_k$ and $\mu_j^{(2)} = \sum_{X_{\mathcal{I}_j \setminus \mathcal{I}_p}} \phi_j \mu_i^{(2)} \prod_{k \in ch(j) \setminus \{i\}} \mu_k$ be factors in $\mathcal{M}_j$, where $j = pa(i)$ and $p = pa(j)$. Since the factors contain only nonnegative values, it follows that $\mu_j^{(2)} \geq \mu_j^{(1)}$. By induction in the nodes of the clique tree, we find that any number $\mu_r^{(1)} \in \mathcal{M}_r$ generated by propagating $\mu_i^{(1)}$ up to the root is dominated by a number $\mu_r^{(2)}$ obtained by propagating $\mu_i^{(2)}$, and therefore (strictly) less than the optimal solution $Z^*$. □

The pruning operation $\text{prune}(\mathcal{M}_i, \sigma, k_i)$ first discards non-extreme and dominated factors from $\mathcal{M}_i$. Albeit accurate, these operations are seldom enough to produce a set $\mathcal{L}_i$ whose cardinality is less than the desired $k_i$. To be able to meet the cardinality constraint, we partition the remaining factors in $\mathcal{M}_i$ (after non-extreme and dominated elements have been removed) in $k_i$ clusters $\mathcal{C}_i^{(1)} \ldots, \mathcal{C}_i^{(k_i)}$, and obtain $\mathcal{L}_i$ by selecting one *representative* factor $\underline{\mu}_i^{(\ell)}$ in each cluster $\mathcal{C}_i^{(\ell)}$. These representatives are valid solutions in that they can be produced from combination of factors from the input sets. Hence, they provide attainable lower bounds for the optimal solution. To account for the (worst-case) errors introduced by the pruning operations we introduce upper bound factors $\sigma(\mu_i)$ for each discarded factor $\mu_i \in \mathcal{C}_i^{(\ell)} \setminus \{\underline{\mu}_i^{(\ell)}\}$. We first discuss how to obtain upper bounds for discarded factors.

Consider a set of factors $\mu_i^{(1)}, \ldots, \mu_i^{(k)}$ which we intend to discard, and let $\overline{\mu}_i$ be a factor such that $\overline{\mu}_i(\mathbf{x}) = \max\{\mu_i^{(1)}(\mathbf{x}), \ldots, \mu_i^{(k)}(\mathbf{x})\}$ for all $\mathbf{x} \in \mathbf{X}_{I_i \cap I_{pa(i)}}$. Then $\overline{\mu}_i \geq \mu_i^{(\ell)}$ for $\ell = 1, \ldots, k$, and it follows from Theorem 3 that any value $\overline{\mu}_r$ obtained by propagating $\overline{\mu}_i$ up to the root is greater than or equal to a solution $\mu_r^{(\ell)}$ obtained by propagating $\mu_i^{(\ell)}$ up to the root, for $\ell = 1, \ldots, k$. Thus, we can use the factor $\overline{\mu}_i$ as an *upper bound* of the factors we wish to discard. We could introduce one upper bound for each discarded factor, but this would cause the propagation of an exponential number of upper bounds (therefore more than the limit $k_i$). On the other extreme, we might produce a single upper bound for all factors discarded from $\mathcal{M}_i$, but this would create too loose a bound. Instead, we generate and propagate one upper bound for each cluster. Let $\underline{\mu}_i^{(\ell)}$ be the representative of a cluster $\mathcal{C}_i^{(\ell)}$. To account for the removal of the elements in the cluster, we update the upper bound $\sigma(\underline{\mu}_i^{(\ell)})$ to be $\max\{\sigma(\mu_i) : \mu_i \in \mathcal{C}_i^{(\ell)}\}$. Figure 2 depicts the pruning of a set $\mathcal{M}_i = \{\mu_i^{(1)}, \mu_i^{(2)}, \mu_i^{(3)}, \mu_i^{(4)}, \mu_i^{(5)}\}$, and the induced upper bounds. Let $\mu_r^{(\ell)}$ be a solution obtained by propagating the representative $\underline{\mu}_i^{(\ell)}$ of cluster $\mathcal{C}_i^{(\ell)}$, and let $\sigma(\mu_r^{(\ell)})$ be the corresponding propagated upper bound. Then it follows that $\mu_r^{(\ell)} \leq Z^* \leq \sigma(\mu_r^{(\ell)})$, where $Z^*$ is the optimal solution of the problem.

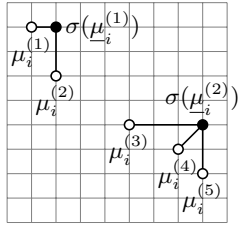

*Figure 2.* A clustering of factors $\mathcal{C}_i^{(1)} = \{\mu_i^{(1)}, \mu_i^{(2)}\}$ and $\mathcal{C}_i^{(2)} = \{\mu_i^{(3)}, \mu_i^{(4)}, \mu_i^{(5)}\}$ with representatives $\underline{\mu}_i^{(1)} = \mu_i^{(1)}$ and $\underline{\mu}_i^{(2)} = \mu_i^{(4)}$, and induced upper bounds $\sigma(\underline{\mu}_i^{(1)})$ and $\sigma(\underline{\mu}_i^{(2)})$.

There still remains to decide how to select good representatives. To this end, we define the following divergence metric $\langle \mu_i^{(1)}, \mu_i^{(2)} \rangle$ that assesses the quality of "representing" a factor $\mu_i^{(1)}$ by a factor $\mu_i^{(2)}$ as $\langle \mu_i^{(1)}, \mu_i^{(2)} \rangle = \max\{\mu_i^{(1)}(\mathbf{x})/\mu_i^{(2)}(\mathbf{x}) : \mathbf{x} \in X\}$. The metric matches the worst-case (multiplicative) error in discarding $\mu_i^{(1)}$ while selecting $\mu_i^{(2)}$ as representative, that is $\mu_i^{(1)} \leq \mu_i^{(2)} \langle \mu_i^{(1)}, \mu_i^{(2)} \rangle$. Note that the divergence is asymmetric, and that it is greater than one if and only if $\mu_i^{(1)}$ is not dominated by $\mu_i^{(2)}$.

Given a set of representatives $\mathcal{V}_i = \{\underline{\mu}_i^{(1)}, \ldots, \underline{\mu}_i^{(k_i)}\}$ in $\mathcal{M}_i$, we assign each factor $\mu_i \in \mathcal{M}_i$ to a cluster $\mathcal{C}_i^{(\ell)}$ such that $\langle \mu_i, \underline{\mu}_i^{(\ell)} \rangle = \min_{l \in [k_i]} \langle \mu_i, \underline{\mu}_i^{(l)} \rangle$. The overall performance of the clustering can be conservatively measured by the largest divergence within a cluster:

$$\epsilon(\mathcal{V}_i) = \max_{\ell \in [k_i]} \max\{\langle \mu_i, \underline{\mu}_i^{(\ell)} \rangle : \mu_i \in \mathcal{C}_i^{(\ell)}\}. \quad (2)$$

Ideally, we would like to find a set $\mathcal{V}_i \subseteq \mathcal{M}_i$ of $k_i$ representatives that obtains the minimum $\epsilon(\mathcal{V}_i)$ over all sets. However, this would add an extra complexity to the computations. Instead, we use a greedy search that at each step attempts to replace a factor in $\mathcal{M}_i \setminus \mathcal{V}_i$ with a factor in $\mathcal{V}_i$ such that $\epsilon(\mathcal{V}_i)$ is decreased.

The following result shows that the the solution found by the algorithm improves monotonically by improving the clusterings at any node of the clique tree.

**Theorem 4.** *The outputs $Z_l$ and $Z_u$ satisfy $Z_u \leq Z_l \prod_{i \in [m]} \epsilon(\mathcal{V}_i)$.*

*Proof.* Consider some inactive node $i$ whose children $j$ are all active, and assume by inductive hypothesis that for any $\mu_j \in \mathcal{L}_j$ it holds that $\sigma(\mu_j) \leq \mu_j e_j$, where $e_j$ is defined as $\epsilon(\mathcal{V}_j) \prod_{k \in de(j)} \epsilon(\mathcal{V}_k)$. Then any $\mu_i \in \mathcal{M}_i$ satisfies $\sigma(\mu_i) = \sum_{\mathbf{X}_{\mathcal{I}_i \setminus \mathcal{I}_p}} \phi_j \prod_{j \in ch(i)} \sigma(\mu_j) \leq \prod_{k \in de(i)} \epsilon(\mathcal{V}_k)[\sum_{\mathbf{X}_{\mathcal{I}_i \setminus \mathcal{I}_p}} \phi_j \prod_{j \in ch(i)} \mu_j] = \mu_i e_i / \epsilon(\mathcal{V}_i)$, where $p = pa(i)$ and $\mu_j \in \mathcal{L}_j$. Let $\underline{\mu}_i$ be the representative of a cluster $\mathcal{C}_i \subseteq \mathcal{M}_i$, with $\sigma(\underline{\mu}_i) = \max\{\mu_i : \mu_i \in \mathcal{C}_i\}$. It follows from (2) that $\sigma(\underline{\mu}_i) \leq \epsilon(\mathcal{V}_i)\underline{\mu}_i$. After the clustering, the new upper bound assigned to $\underline{\mu}_i$ is (by design) given by $\overline{\mu}_i = \max\{\sigma(\mu_i) : \mu_i \in \mathcal{C}_i\}$, which satisfies $\overline{\mu}_i \leq \sigma(\underline{\mu}_i) e_i / \epsilon(\mathcal{V}_i) \leq e_i \underline{\mu}_i$. □

The above result guarantees that the algorithm finds lower and upper bounds whose ratio is not worse than the product of the clustering quality measures $\prod_{i \in [m]} \epsilon(\mathcal{V}_i)$. The quality of each cluster $\epsilon(\mathcal{V}_i)$ can be improved by increasing the maximum allowed number of elements $k_i$ in the set. Since each set cannot have more than $|\mathcal{K}_1| \cdots |\mathcal{K}_m|$ elements, the algorithm is guaranteed to converge to the optimum in finite time. In fact, each maximum set size $k_i$ needs only to be as high as the number of extrema and non-dominated factors in $\mathcal{M}_i$, since these are shown to lead to exact computations. These remarks lead naturally to the anytime algorithm we present in the next section.

## 6. Anytime Inference

An anytime algorithm is a procedure that can be interrupted at any time with a meaningful solution whose quality is a monotonic function of runtime. Hence, anytime algorithms allow a trade-off between computation time and quality of solutions.

We can easily transform FACTOR-SET-ELIMINATION into an anytime algorithm that continuously improve the lower and upper bounds by increasing the maximum set cardinalities $k_1, \ldots, k_m$. The procedure is described in Algorithm 3. The anytime algorithm starts by running FACTOR-SET-ELIMINATION with all maxi-mum set cardinalities $k_1^{(0)}, \ldots, k_m^{(0)}$ set to one. This produces an arbitrary (but feasible) lower bound $Z_l^{(0)}$, and an upper bound $Z_u^{(0)}$ that matches the value returned by FACTOR-MAX-ELIMINATION. Then, for each time step, the algorithm increases the maximum set cardinality $k_i$ of the node $i$ with poorest clustering quality $\epsilon(\mathcal{V}_i)$ by a given constant $c$. In principle, even if we improve the clustering quality we might obtain a worse solution, as the metric that evaluates clustering quality optimizes worst case. This can be circumvented by enlarging each set $\mathcal{L}_i$ incrementally.

**Algorithm 3** ANYTIME-INFERENCE
**Require:** A clique tree $\mathcal{T}$ over sets $\mathcal{K}_1, \ldots, \mathcal{K}_m$ and integer $c$
1: let $k_1^{(0)} = 1, \ldots, k_m^{(0)} = 1$, $Z_l^{(0)} = 0$ and $Z_u^{(0)} = 1$
2: set $t \leftarrow 0$
3: **while** $Z_l^{(t)} < Z_u^{(t)}$ and not interrupted **do**
4:    find the node $i$ with highest $\epsilon(\mathcal{V}_i)$
5:    run FACTOR-SET-ELIMINATION with $k_1^{(t)}, \ldots, k_m^{(t)}$ and let $(Z_l, Z_u)$ be its output
6:    set $Z_l^{(t+1)} = \max\{Z_l, Z_l^{(t)}\}$, $Z_u^{(t+1)} = \min\{Z_u, Z_u^{(t)}\}$ and $k_i^{(t+1)} = k_i^{(t)} + c$, $i = 1, \ldots, m$
7:    set $t \leftarrow t + 1$
8: **end while**

| Net | $n$ | $d$ | SI | AFSE | $Z/Z^*$ |
|---|---|---|---|---|---|
| Insurance | 27 | 2 | 0.2s | 0.9s | 1 |
| Alarm | 37 | 12 | 0.1s | 0.2s | 1 |
| Barley | 48 | 10 | 10s | >1h | 0.1 |
| Hailfinder | 56 | 17 | 0.5s | 1.4s | 1 |
| Pigs | 441 | 145 | 6m | 5m | 1 |
| KS-3-20 | 42 | 20 | 2.2s | 0.1s | 1 |
| KS-3-50 | 102 | 50 | >1h | 0.4s | 1 |
| KS-3-100 | 202 | 100 | >1h | 11m | 1 |
| Grid-4-10-2 | 80 | 40 | >1h | 7m | 0.96 |
| Grid-4-25-2 | 200 | 100 | >1h | 22m | 0.73 |
| Grid-4-30-2 | 240 | 120 | >5h | 2.6h | 0.55 |
| Grid-6-6-1 | 36 | 20 | 1.1s | 0.1s | 1 |
| Grid-10-10-1 | 100 | 36 | 1.8s | 7s | 1 |
| Grid-16-16-1 | 256 | 60 | 48s | 12m | 1 |
| Grid-18-18-1 | 324 | 68 | – | 2.9h | – |

Table 1. Performance of Anytime Factor-Set-Elimination algorithm (AFSE) and SamIam (SI) on real and synthetic models.

## 7. Experiments

We performed experiments with three groups of graphical models, which range from simple to very challenging problems. The first group, which appears in the top five lines of Table 1, consists of benchmark Bayesian networks used in real applications.[2] In these networks, the MAP inference asks for optimum assignments of the root nodes given some evidence on every leaf. This creates MAP problems where every variable in the network is relevant to the solution and obtaining an exact solution by FACTOR-MAX-ELIMINATION would take time (at least) exponential in the number of decision variables. The second group (lines 6–8 of the table) contains graphical models designed to solve multiple knapsack problems with three bags and varying number of items (20, 50 and 100). The graphs are structured in a chain of latent variables with root decision nodes as parents. Besides the importance of the multiple knapsack itself, this group allows us to evaluate the performance of the methods when the search space is large but the treewidth is low. Finally, the third group (last seven lines of the table) consists of grid-structured graphical models whose pa-rameters were uniformly sampled. Each Grid-$x$-$y$-$z$ model contains $x$ rows, $y$ columns and $z$ planes. For $z = 2$, variables are quaternary and the grid has two planes: one is the grid itself and the other is formed by decision variables that are linked to grid variables in a one-to-one correspondence; for $z = 1$, the models are usual planar binary grids, with all border variables chosen as decision variables. These experiments allow us to better evaluate how the performance is affected by the treewidth and the size of the search space.

---

[2]At the time of this submission, they were available at *http://www.cs.huji.ac.il/site/labs/compbio/Repository/*.

We compare our algorithm against SamIam's implementation of the systematic search algorithm of Park & Darwiche (2003), which we refer to as SI. We chose SI because (i) it is a state-of-the-art algorithm, (ii) its implementation is publicly available, (iii) it is an anytime procedure, and (iv) it returns feasible solutions.

Table 1 shows the results of the experiments, comparing the proposed method (named AFSE for short) and SI. The table presents names, total number of variables, number of decision variables, amount of time that SI and AFSE, respectively, spent to solve the instances, and errors of the obtained solution (in case one of the methods was unable to solve the instances in a reasonable amount of time and memory). The error corresponds to the worst of the two methods and it was obtained by calculating the ratio of the returned value and the optimum (the worst of the two methods can be identified in the columns corresponding to the time they spent, indicated by a ">$t$", where $t$ is the time-limit used for the given method).

Some results from Table 1 deserve an additional discussion. Firstly, models in the second group and the two-plane grids of the third group of experiments in-dicate that AFSE is by far faster when treewidth is small. Still, SI was able to find the best solution in the models of the second group (even though it was not aware of it, so the search have not stopped), but clearly degrades in the two-plane grids, as can be seen in the error column of the table, which reaches 55% in Grid-4-30-2. This means that not only the algorithm did not finish but the best solution found was far from the optimum. Such situation justifies the use of methods that can provide anytime lower and upper bounds for the solution. Secondly, AFSE performed similarly to SI in (real) Bayesian networks, with the largest differences in the Barley and Pigs networks (the former favorable to SI, the latter favorable to AFSE). We see on the squared grids of the third group that SI can handle better the increase of treewidth, indeed a known characteristic of SI. The exception is Grid-18-18-1, where SI exhausted the 8 GB of memory granted without being able to produce a (candidate) solution. Finally, the time-accuracy trade-off of the algorithms can be seen in Figure 3, which shows the accuracy of AFSE and SI on models Grid-4-30-2 and Grid-4-25-2 as a function of time. Lower and upper bounds converge to the optimal solution, and while SI starts with a better lower bound, it gets stuck in the search and does not converge within the allowed time.

## 8. Conclusion

We present a new anytime algorithm for the marginal MAP assignment problem in graphical models. We show theoretically that the algorithm produces feasible solutions whose quality are a function of the amount of computational resources granted. The convergence and error bounds are analyzed.

By performing experiments with real and synthetic graphical models, we show that the proposed algorithm is competitive against the systematic search of Park & Darwiche (2003). In particular, our algorithm compares favorably when the problems exhibit moderate treewidth but large search space. Unfortunately, as the treewidth increases, the bounds returned by the algorithm become too loose. This could be mitigated by decomposing the propagated factors into smaller domains, as in the work of Meek & Wexler (2011).

Understanding how the numerical parameters of the input affect the complexity of the algorithm is an important question that remains open. Finally, in the spirit of the result by de Campos (2011) it is possible to show that the anytime algorithm is also a fully polynomial-time approximation scheme for graphical models if we assume that both the treewidth and the number of values a variable can assume are bounded.

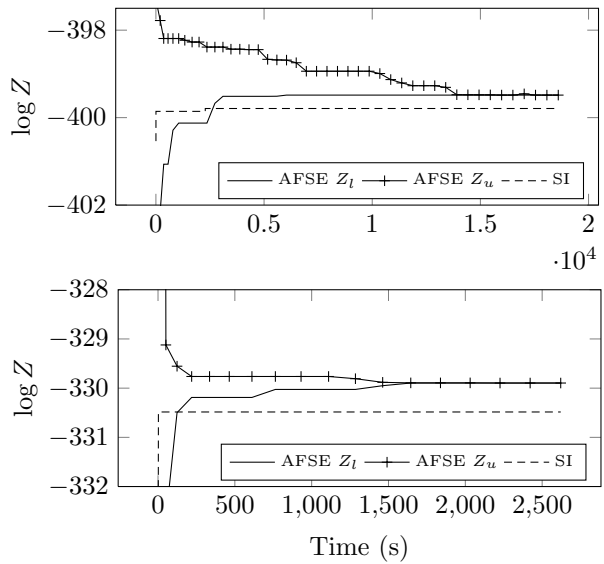

Figure 3. Quality of the solutions of AFSE and SI on the Grid-4-30-2 and Grid-4-25-2 models by running time.

## Acknowledgments

This work was partially supported by the Swiss NSF grants no. 200020_134759/1 and 200020_132252.